%% file: root.tex
\documentclass[letterpaper, 10pt, conference]{ieeeconf}  

\IEEEoverridecommandlockouts                              

\input{packages_config.tex}

\author{Levent Ögretmen, Mo Chen, Phillip Pitschi, and Boris Lohmann
	\thanks{
		All authors are with the Chair of Automatic Control, Department of Mechanical Engineering, TUM School of Engineering and Design, Technical University of Munich, 85748 Garching, Germany
		{\tt
			\{%
				\href{mailto:levent.oegretmen@tum.de}{levent.oegretmen}, %
				\href{mailto:ge64mah@tum.de}{ge64mah}, %
				\href{mailto:phillip.pitschi@tum.de}{phillip.pitschi}, %
				\href{mailto:lohmann@tum.de}{lohmann}%
			\}@tum.de
		}
	}%
}

\title{\LARGE \bf
	Trajectory Planning Using Reinforcement Learning for Interactive Overtaking Maneuvers in Autonomous Racing Scenarios
}

\newcommand\copyrighttext{%
	\footnotesize \textcopyright 2024 IEEE. Personal use of this material is permitted. Permission from IEEE must be obtained for all other uses, in any current or future media, including reprinting/republishing this material for advertising or promotional purposes, creating new collective works, for resale or redistribution to servers or lists, or reuse of any copyrighted component of this work in other works.
}

\newcommand\copyrightnotice{%
	\begin{tikzpicture}[remember picture,overlay]
		\node[anchor=south,yshift=5pt] at (current page.south) {\fbox{\parbox{\dimexpr\textwidth-\fboxsep-\fboxrule\relax}{\copyrighttext}}};
	\end{tikzpicture}%
}

\begin{document}
	
	\maketitle
	\backgroundsetup{opacity=1, scale=1, color=black, angle=0, contents={\copyrightnotice}} 
	\BgThispage
	\thispagestyle{empty}
	\pagestyle{empty}
	
	\input{chapters/abstract.tex}
	\input{chapters/introduction.tex}
	\input{chapters/preliminaries.tex}
	\input{chapters/blocking_scenario.tex}

	\input{chapters/methodology.tex}
	\input{chapters/results_discussion.tex}
	\input{chapters/conclusion_outlook.tex}
	
	\bibliographystyle{IEEEtran}
	\bibliography{IEEEabrv, references}

\end{document}

%% file: packages_config.tex
\usepackage{cite}
\usepackage{amsmath,amssymb,amsfonts}
\usepackage{algorithmic}
\usepackage{graphicx}
\usepackage{textcomp}
\usepackage{xcolor}

\usepackage{acro}			              
\usepackage{bm}			                  
\usepackage{siunitx}		              
\usepackage{hyperref}   	              
\usepackage{booktabs}   	              
\usepackage{tikz, pgfplots, tikz-3dplot}  
\usepackage{xfrac}						  
\usepackage{dsfont}						  
\usepackage[pages=some]{background} 	  

\pgfplotsset{compat=newest}

\usetikzlibrary{positioning}  
\usetikzlibrary{arrows.meta}  
\usetikzlibrary{calc}         
\usetikzlibrary{decorations.markings}
\usepgfplotslibrary{groupplots}

\newlength\figH
\newlength\figW
\newlength\horiDis

\definecolor{TUMBlue}{HTML}{0065BD}            
\definecolor{TUMBlack}{HTML}{000000}           
\definecolor{TUMWhite}{HTML}{ffffff}           
\definecolor{TUMBlueLight}{HTML}{005293}       
\definecolor{TUMBlueDark}{HTML}{003359}        
\definecolor{TUMGrayLight}{HTML}{CCCCCC}       
\definecolor{TUMGray}{HTML}{808080}            
\definecolor{TUMGrayDark}{HTML}{333333}        
\definecolor{TUMAccGray}{HTML}{DAD7CB}         
\definecolor{TUMAccGreen}{HTML}{A2AD00}        
\definecolor{TUMAccOrange}{HTML}{E37222}       
\definecolor{TUMAccBlueDark}{HTML}{64A0C8}     
\definecolor{TUMAccBlueLight}{HTML}{98C6EA}    

\hypersetup{
	pdffitwindow    = false,
	pdfstartview    = FitV,
	pdfnewwindow    = false,
	colorlinks      = true,
	linkcolor       = TUMBlack,
	citecolor       = TUMBlack,
	pdfborder       = {0 0 0},
	urlcolor        = TUMBlack,
}

\DeclareAcronym{CH}{
	short=CH,
	long=constant heading
}

\DeclareAcronym{CLP}{
	short=CLP,
	long=constant lateral position
}

\DeclareAcronym{IAC}{
	short=IAC,
	long=Indy Autonomous Challenge
}

\DeclareAcronym{MDP}{
	short=MDP,
	long=Markov Decision Process
}

\DeclareAcronym{PPO}{
	short=PPO,
	long=proximal policy optimization
}

\DeclareAcronym{RL}{
	short=RL,
	long=Reinforcement Learning
}

\DeclareAcronym{SL}{
	short=SL,
	long=safety layer
}


%% file: chapters/abstract.tex
\begin{abstract}
Conventional trajectory planning approaches for autonomous racing are based on the sequential execution of prediction of the opposing vehicles and subsequent trajectory planning for the ego vehicle. If the opposing vehicles do not react to the ego vehicle, they can be predicted accurately. However, if there is interaction between the vehicles, the prediction loses its validity. For high interaction, instead of a planning approach that reacts exclusively to the fixed prediction, a trajectory planning approach is required that incorporates the interaction with the opposing vehicles. This paper demonstrates the limitations of a widely used conventional sampling-based approach within a highly interactive blocking scenario. We show that high success rates are achieved for less aggressive blocking behavior but that the collision rate increases with more significant interaction. We further propose a novel \ac{RL}-based trajectory planning approach for racing that explicitly exploits the interaction with the opposing vehicle without requiring a prediction. In contrast to the conventional approach, the \ac{RL}-based approach achieves high success rates even for aggressive blocking behavior. Furthermore, we propose a novel \ac{SL} that intervenes when the trajectory generated by the \ac{RL}-based approach is infeasible. In that event, the \ac{SL} generates a sub-optimal but feasible trajectory, avoiding termination of the scenario due to a not found valid solution.
\end{abstract}

%% file: chapters/introduction.tex
\section{Introduction}
\label{sec:introduction}
Conventional trajectory planning for autonomous racing consists of a sequential execution of prediction and planning \cite{Betz2022}. Initially, the movement of the opposing vehicles is predicted based on specific assumptions. Subsequently, the trajectory of the ego vehicle is planned, avoiding collisions with the assumed prediction. This sequential approach is suitable if the behavior of the opposing vehicles is independent of the ego vehicle. However, if an interaction exists between the vehicles, i.e., the opposing vehicles react to the behavior of the ego vehicle, no correct prediction can be determined. 

An example of such an interactive scenario is an overtaking attempt, which the opposing vehicle actively tries to prevent by blocking. If the blocking vehicle is predicted to maintain a constant distance from a track boundary and the track is sufficiently wide, an overtaking maneuver is initiated. After starting to overtake, however, the blocking vehicle reacts immediately and thus deviates from the previous prediction. Overall, no valid prediction can be stated since the planning depends on the prediction, and in turn, the behavior of the opposing vehicle depends on the planned trajectory. This simple example encourages using more advanced interaction-aware planning approaches that avoid the sequential execution of prediction and planning. A promising approach for this is \ac{RL}, which allows for learning complex policies in high-dimensional environments with non-linear interaction and dynamics \cite{Kiran2022}.

This paper examines the interactive scenario described above, in which a blocking vehicle is to be overtaken. In this scenario, we evaluate an existing, widely used conventional planning approach and demonstrate its limitations. Further, we propose a novel \ac{RL}-based planning approach that explicitly exploits the interaction with the blocking vehicle and evaluate its performance and generalization ability. Additionally, we introduce an \ac{SL} that counters infeasible trajectories caused by a lack of generalization of the \ac{RL}-based approach.

\subsection{Related Work}
\label{subsec:related_work}
Conventional trajectory planning approaches require a prediction of the opposing vehicles, which is generated based on underlying assumptions. In current autonomous racing series, such as the \ac{IAC}\footnote{\url{https://indyautonomouschallenge.com}}, with strict racing rules, often a constant speed and distance to the track boundary is assumed by the teams, as in \cite{Betz2023, Jung2023, Raji2024}. For the subsequent trajectory planning, there are sampling-based, graph-based, and optimization-based approaches \cite{Betz2022}. A commonly used sampling-based approach proposed in \cite{Werling2010} for traffic scenarios is to generate a set of jerk-optimal trajectories by sampling the end state. The trajectories are checked for feasibility, and the optimal valid one is selected according to a pre-defined cost function. The described concept is applied similarly in \cite{Jung2023, Raji2024} for racing on oval tracks. The approach is extended in \cite{Oegretmen2024b} to complex racing circuits with further consideration of the three-dimensional effects of the track. In \cite{Oegretmen2022}, the jerk-optimal trajectories are used within a graph-based approach to connect the vehicle state with an offline computed spatio-temporal graph.

In addition to the conventional methods, there exist \ac{RL}-based motion planning methods, which are summarized in \cite{Aradi2022}. In these methods, a general distinction is made between the action space, i.e., the level of control over the vehicle. On one side, there are low-level approaches in which the control inputs of the vehicle, e.g., acceleration and steering angle, are selected directly. On the other side, there are high-level approaches in which a maneuver, such as lane change left/right, keep lane, accelerate, or brake, is selected from a discrete finite set, and a fixed underlying movement is executed. The low-level approaches have been applied extensively for solo racing \cite{Jaritz2018, Guckiran2019, Fuchs2021, Evans2023}, which naturally does not include interaction. A few studies have also examined overtaking in a multi-vehicle racing scenario \cite{Loiacono2010, Song2021, Wurman2022}. In \cite{Loiacono2010} and \cite{Song2021}, the opposing vehicles follow either a fixed or random trajectory without interaction. In \cite{Wurman2022}, the opponent vehicles are controlled by built-in game agents and older versions of the \ac{RL} agent, with no information on their interaction. The direct control of the vehicle using these low-level methods provides the greatest possible freedom of maneuvers but can make training more complex, as the entire dynamics of the vehicle must be learned. In addition, an application on a real vehicle is safety-critical, as the directly applied inputs can lead to unsafe behavior due to modeling inaccuracies between the training model and the real vehicle. In contrast, the high-level approaches operate on a more abstract level, making them less safety-critical for real-world application. However, they are less suitable for the racing scenario since a restriction to a discrete, finite set of maneuvers does not adequately represent all necessary maneuvers in the mainly unconstrained racing scenario. Such approaches are often used in highway scenarios instead, as the definition of meaningful maneuvers is easier due to the structured environment with lanes.

Instead of the high- or low-level approaches mentioned above, it is also possible to plan a trajectory directly. Compared to the low-level approaches, trajectory generation is preferable from a safety point of view, as the mapping to control inputs is performed by a conventional controller that can be tuned to the specific vehicle being used. Compared to the high-level approaches, there is no need to define a set of discrete high-level maneuvers with fixed underlying trajectories, allowing for more diverse trajectories. Direct trajectory generation has been used for traffic scenarios, but to the authors' knowledge, there is no application for racing. The trajectory is usually generated by utilizing an underlying structure to reduce the complexity of the training process. The authors of \cite{Trauth2024} utilize the same procedure as in \cite{Werling2010} to generate a set of jerk-optimal trajectories by sampling the end state and use the weighting parameters of the cost function as the action space. While this approach is sufficient for traffic scenarios, it is less suitable for racing as it is limited to the generated set of trajectories, preventing the realization of more complex maneuvers. In \cite{Mirchevska2023}, no sampling is performed, and instead, the end state of the jerk-optimal trajectory is used as the action space, which allows for more diverse maneuvers. If the end state chosen by the \ac{RL} agent leads to an infeasible trajectory, the scenario is terminated immediately unsuccessfully.

For completeness, it should be mentioned that in addition to the \ac{RL}-based approaches described here, game-theoretical approaches are also used for interactive planning. However, in contrast to \ac{RL}-based approaches, they require increased online computing times, as a time-consuming game has to be solved in each execution step. In \cite{Liniger2020}, a two-player blocking game is examined, and a bi-matrix game is solved based on sampled trajectory candidates. In \cite{Wang2021}, the trajectories of the individual vehicles are optimized iteratively, while the trajectories of the opposing vehicles are fixed. 

\subsection{Contribution}
\label{subsec:contribution}
In contrast to existing literature, we focus in this work on interactive overtaking for autonomous racing and compare a conventional trajectory planning approach to a novel \ac{RL}-based approach. The main contributions are summarized as follows:

\begin{itemize}
	\item We introduce a highly interactive blocking scenario with a rule-based blocking agent, evaluate a commonly used conventional sampling-based trajectory planning approach in this setting, and show its limitations.
	\item We propose an \ac{RL}-based trajectory planning approach for the introduced blocking scenario. In contrast to existing approaches for racing with a low-level action space, the trajectory is generated directly, avoiding the drawbacks described in Section~\ref{subsec:related_work}. The approach is evaluated in the blocking scenario and compared with the conventional approach.
	\item For the \ac{RL}-based approach, we introduce an \ac{SL} that intervenes if the generated trajectory is infeasible. In this case, a set of trajectories is generated, and the valid, most similar trajectory is selected. Thus, in contrast to existing approaches, infeasibility does not terminate the scenario, increasing the success rate.
\end{itemize}

The remainder of this paper is structured as follows: Section~\ref{sec:rl_preliminaries} contains introductory preliminary remarks on \ac{RL}. The interactive blocking scenario considered here, including the behavior of the blocking vehicle, is presented in Section~\ref{sec:blocking_scenario}. Section~\ref{sec:methodology} presents the conventional sampling-based and the proposed \ac{RL}-based trajectory planning approach with the \ac{SL}. The evaluation of both approaches in the blocking scenario is carried out in Section~\ref{sec:results_discussion}, followed by a summary and an outlook for further research in Section~\ref{sec:conclusion_outlook}.

%% file: chapters/preliminaries.tex
\section{Reinforcement Learning Preliminaries}
\label{sec:rl_preliminaries}
\ac{RL} is a method where an agent learns by interacting with an environment in a try-and-error style. A \ac{MDP} mathematically formalizes the \ac{RL} problem with the tuple $\left( \mathfrak{s}, \mathfrak{a}, P(s' \vert s,a), R(s,a), \gamma \right)$. $\mathfrak{s}$ and $\mathfrak{a}$ are the state space and action space, respectively. When being in state $s \in \mathfrak{s}$ and choosing action $a \in \mathfrak{a}$, the state transition function $P(s' \vert s,a)$ describes the probability of ending up in the state $s' \in \mathfrak{s}$. $R(s,a)$ is the expected reward the agent receives for this transition. The \ac{RL} agent chooses the action $a$ according to a policy $\pi(a \vert s)$. The goal of \ac{RL} is to find a policy that maximizes the return $G_t = \sum_{k=t+1}^{T_\mathrm{ep}} \gamma^{k-t-1} \, R_k$, which is the discounted sum of all future rewards from the current time step $t$ to the terminal time step $T_\mathrm{ep}$ of the episode. The discount factor $\gamma \in [\num{0},\num{1})$ scales how much the future rewards are worth compared to immediate rewards. The value function $v_{\pi}(s) = \mathbb{E}_{\pi}[G_t \vert s]$ for a given policy is defined as the expected return when being in state $s$ and thereafter following the policy $\pi(a \vert s)$. \cite{Sutton2018}

\ac{RL} algorithms reach their goal by learning the value function (value-based) or a policy directly (policy-based). Methods that use both are called actor-critic algorithms. The actor is the policy $\pi(a \vert s)$ that acts on the environment. The critic is the value function $v_{\pi}(s)$ that evaluates the reaction of the environment. One widely used representative of this method is the \ac{PPO} algorithm \cite{Schulman2017}. It improves the general actor-critic approach by constraining the size of the policy updates.

\ac{RL} algorithms are often combined with curriculum learning to improve the results. This method divides the learning procedure into multiple tasks with usually increasing complexity. The \ac{RL} algorithm learns from the samples generated from one task and transfers the gained knowledge to the next task until it arrives at the final task. \cite{Bengio2009, Narvekar2020}

%% file: chapters/blocking_scenario.tex
\section{Blocking Scenario}
\label{sec:blocking_scenario}
The blocking scenario under consideration is introduced with the used race track in Section~\ref{subsec:track}, the general scenario description in Section~\ref{subsec:goal}, and the behavior of the blocking vehicle in Section~\ref{subsec:blocking_vehicle}.

\subsection{Race Track}
\label{subsec:track}
The blocking scenario is carried out on a straight track without inclination or banking. The track is \SI{1500}{\meter} long and \SI{15}{\meter} wide. The center line of the track is used as the reference line with which the curvilinear coordinates $s$ and $n$ can be defined. While $s$ corresponds to the progress along the reference line, $n$ is the lateral offset along the corresponding normal vector. Although the orientation $\theta_\mathrm{r}(s)$ and curvature $\kappa_\mathrm{r}(s)$ of the reference line are \num{0} for straight tracks, they are accounted for in the following expressions for reasons of generality. Finally, we denote the distance between the reference line and the left and right track boundary with $n_{\mathrm{l}} = \SI{7.5}{\meter}$ and $n_{\mathrm{r}} = \SI{7.5}{\meter}$.

\subsection{Scenario Description and Goal}
\label{subsec:goal}
A scenario consisting of two vehicles is considered on the introduced track, as illustrated in Fig.~\ref{fig:scenario}. The overtaking vehicle (variables denoted with index $\mathrm{o}$) is initialized on the reference line at the beginning of the track with $s_\mathrm{o, init} = \SI{0}{\meter}$ and $n_\mathrm{o, init} = \SI{0}{\meter}$. The opposing blocking vehicle (variables denoted with index $\mathrm{b}$) is initialized with a longitudinal gap at $s_\mathrm{b, init}>0$ and a lateral offset $n_\mathrm{b, init}$. Both vehicles are initialized with the same velocity $v_\mathrm{init}>0$.

\input{figures/scenario.tex}

The blocking vehicle has a fixed rule-based blocking behavior described in Section~\ref{subsec:blocking_vehicle}. For the overtaking vehicle, however, trajectories are planned using the conventional or \ac{RL}-based approach, both introduced in Section~\ref{sec:methodology}, and it is assumed that the planned trajectories are followed exactly. The scenario ends successfully if the blocking vehicle is overtaken. However, if a planned trajectory is not feasible, a collision occurs between the vehicles, or no overtaking maneuver could be carried out by the end of the race track, the scenario ends unsuccessfully.

\subsection{Blocking Vehicle}
\label{subsec:blocking_vehicle}
A kinematic bicycle model is used to specify the blocking vehicle kinematics. Omitting the time argument $t$ for brevity and noting the derivative with respect to time as $\frac{\mathrm{d}\square}{\mathrm{d} t} = \dot{\square}$, it follows:
\begin{equation}
	\begin{split}
		\dot{s}_\mathrm{b} &= v_\mathrm{b} \cdot \cos(\chi_\mathrm{b}) \cdot \frac{1}{1 - n_\mathrm{b} \cdot \kappa_\mathrm{r}(s_\mathrm{b})} \\
		\dot{n}_\mathrm{b} &= v_\mathrm{b} \cdot \sin(\chi_\mathrm{b}) \\
		\dot{\chi}_\mathrm{b} &= \frac{v_\mathrm{b}}{l_{\mathrm{r}}} \cdot \sin(\beta_\mathrm{b}) - v_\mathrm{b} \cdot \cos(\chi_\mathrm{b}) \cdot \frac{\kappa_\mathrm{r}(s_\mathrm{b})}{1 - n_\mathrm{b} \cdot \kappa_\mathrm{r}(s_\mathrm{b})} \\
		\dot{v}_\mathrm{b} &= a_{\mathrm{b}} \\
		\dot{\delta}_\mathrm{b} &= \omega_{\mathrm{b}}
	\end{split}
	\label{eq:system_dynamics}
\end{equation}
with the body slip angle $\beta_\mathrm{b} = \arctan\left( \frac{l_{\mathrm{r}}}{l_{\mathrm{r}} + l_{\mathrm{f}}} \cdot \tan(\delta_\mathrm{b}) \right)$. The state vector $\bm{x} = \begin{bmatrix} s_\mathrm{b} & n_\mathrm{b} & \chi_\mathrm{b} & v_\mathrm{b} & \delta_\mathrm{b} \end{bmatrix}^\intercal$ contains the position of the rear axle in curvilinear coordinates $s_\mathrm{b}$ and $n_\mathrm{b}$, the relative orientation to the reference line $\chi_\mathrm{b}$, the absolute velocity $v_\mathrm{b}$, and the steering angle $\delta_\mathrm{b}$. The input vector $\bm{u} = \begin{bmatrix} \omega_{\mathrm{b}} & a_{\mathrm{b}} \end{bmatrix}^\intercal$ consists of the steering rate $\omega_{\mathrm{b}}$ and the longitudinal acceleration $a_{\mathrm{b}}$. The distance from the center of gravity to the rear and front axle is denoted as $l_{\mathrm{r}}$ and $l_{\mathrm{f}}$ respectively. We further constrain the steering angle $|\delta_{\mathrm{b}}(t)| \leq \delta_{\mathrm{b},\max}$ and the steering rate $|\omega_{\mathrm{b}}(t)| \leq \omega_{\mathrm{b},\max}$.

\setlength{\figH}{0.55\columnwidth}
\setlength{\figW}{1.0\columnwidth}
\begin{figure}[t]
	\centering
	\small
	\input{figures/step_responses.tex}
	\caption{Step responses for different values of $s_{\mathrm{d}}$ with $k_{\mathrm{p}}=\num{0.05}$, $k_{\mathrm{d}}=\num{0.6}$, and $k_{\mathrm{n}}=\num{1.0}$. The aggressiveness of the blocking maneuver increases with lower values of $s_{\mathrm{d}}$.}
	\label{fig:step_responses}
\end{figure}

For the longitudinal motion, we select $a_{\mathrm{b}}(t) = \SI{0}{\meter\per\second\squared}$, meaning that a constant speed is maintained. For the lateral motion, we specify the steering rate $\omega_{\mathrm{b}}(t)$ based on a PD controller with parameters $k_{\mathrm{p}}$ and $k_{\mathrm{d}}$:
\begin{equation}
	\begin{split}
		\omega_{\mathrm{b}}(t) &= k_{\mathrm{p}} \cdot e(t) + k_{\mathrm{d}} \cdot \dot{e}(t) \\
		e(t) &= \chi_{\mathrm{d}}(t) - \chi_{\mathrm{b}}(t)
	\end{split}
\end{equation}
that minimizes the error $e(t)$ between the actual vehicle orientation $\chi_{\mathrm{b}}(t)$ and a desired orientation $\chi_{\mathrm{d}}(t)$. The desired orientation $\chi_{\mathrm{d}}(t)$ is calculated using \eqref{eq:chi_desired} based on a desired lateral offset $\Delta \tilde{n}(t)$ and a lookahead distance parameter $s_{\mathrm{d}}$. In addition to the lateral distance $n_{\mathrm{o}}(t) - n_{\mathrm{b}}(t)$ between the overtaking and blocking vehicle, $\Delta \tilde{n}(t)$ also considers the lateral velocity, incorporating the parameter $k_{\mathrm{n}}$.
\begin{equation}
	\begin{split}
		\chi_{\mathrm{d}}(t) &= \arctan\left( \Delta \tilde{n}(t) \cdot s_{\mathrm{d}}^{-1} \right) \\
		\Delta \tilde{n}(t) &= n_{\mathrm{o}}(t) - n_{\mathrm{b}}(t) + k_{\mathrm{n}} \cdot \left( \dot{n}_{\mathrm{o}}(t) - \dot{n}_{\mathrm{b}}(t) \right)
	\end{split}
	\label{eq:chi_desired}
\end{equation}
In particular, the lookahead distance parameter $s_{\mathrm{d}}$ in the control law has a decisive influence on the aggressiveness of the blocking vehicle. The lower $s_{\mathrm{d}}$ is selected, the greater the magnitude of $\omega_{\mathrm{b}}$ and the higher the aggressiveness. The step responses for different values of $s_{\mathrm{d}}$ are shown in Fig.~\ref{fig:step_responses}.

%% file: figures/scenario.tex

\def\trackwidth{0.25*\columnwidth}
\def\tracklength{0.86*\columnwidth}


\def\dimensiondistance{0.3cm}

\def\vehiclewidth{0.2*\trackwidth}
\def\vehiclelength{0.4*\trackwidth}

\def\xovertaker{0.0}
\def\yovertaker{0.0}

\def\xblocker{0.5*\tracklength}
\def\yblocker{-0.33*\trackwidth}

\def\cslength{0.35*\trackwidth}

\begin{figure}[t]
	\centering
	\begin{tikzpicture}
		
		\draw[dashed, thick] ({0},{0}) -- ({\tracklength},{0}) node[above, pos=0.8] {Reference line};
		
		\draw[very thick] ({0},{\trackwidth/2}) -- ({\tracklength},{\trackwidth/2});
		
		\draw[very thick] ({0},{-\trackwidth/2}) -- ({\tracklength},{-\trackwidth/2});
		
		\draw[Triangle-Triangle] ({0}, {-\dimensiondistance-\trackwidth/2}) -- ({\tracklength},{-\trackwidth/2-\dimensiondistance}) node[midway,below] {\SI{1500}{\meter}};
		\draw[Triangle-Triangle] ({\tracklength+\dimensiondistance},{-\trackwidth/2}) -- ({\tracklength+\dimensiondistance},{\trackwidth/2}) node[midway,below,rotate=90] {\SI{15}{\meter}};
		
		\draw[draw=TUMBlack, fill=TUMBlue] ({\xovertaker-\vehiclelength/2},{\yovertaker-\vehiclewidth/2}) rectangle ++(\vehiclelength,\vehiclewidth);
		
		\draw[draw=TUMBlack, fill=TUMAccGreen] ({\xblocker-\vehiclelength/2},{\yblocker-\vehiclewidth/2}) rectangle ++(\vehiclelength,\vehiclewidth);
		
		\draw[-Triangle, very thick] ({0}, {0}) -- ({\cslength},{0}) node[above right, pos=0.65] {$s$};
		\draw[-Triangle, very thick] ({0}, {0}) -- ({0},{\cslength}) node[right, pos=0.9] {$n$};
		
		\draw[-] ({0}, {0}) -- ({0},{\yblocker*1.15});
		\draw[Triangle-Triangle] ({0}, {\yblocker}) -- ({\xblocker},{\yblocker}) node[above, pos=0.5, yshift=-0.1cm] {$s_\mathrm{b, init}$};
		\draw[Triangle-Triangle] ({\xblocker}, {0}) -- ({\xblocker},{\yblocker}) node[right, pos=0.5] {$n_\mathrm{b, init}$};

		
		
		
		
		
	\end{tikzpicture}
	\caption{Initialization of the blocking scenario with the overtaking vehicle (blue) at position $(s_\mathrm{o, init} = \SI{0}{\meter}, n_\mathrm{o, init} = \SI{0}{\meter})$ and the blocking vehicle (green) at $(s_\mathrm{b, init}, n_\mathrm{b, init})$. The reference line of the considered straight race track (not to scale) is depicted as a dashed black line.}
	\label{fig:scenario}
\end{figure}

%% file: figures/step_responses.tex
\begin{tikzpicture}

\definecolor{crimson2143940}{RGB}{214,39,40}
\definecolor{darkgray176}{RGB}{176,176,176}
\definecolor{darkorange25512714}{RGB}{255,127,14}
\definecolor{forestgreen4416044}{RGB}{44,160,44}
\definecolor{lightgray204}{RGB}{204,204,204}
\definecolor{mediumpurple148103189}{RGB}{148,103,189}
\definecolor{sienna1408675}{RGB}{140,86,75}
\definecolor{steelblue31119180}{RGB}{31,119,180}

\begin{axis}[
height=\figH,
legend cell align={left},
legend columns=2,
legend style={
  fill opacity=0.8,
  draw opacity=1,
  text opacity=1,
  at={(0.97,0.03)},
  anchor=south east,
  draw=lightgray204
},
tick align=outside,
tick pos=left,
width=\figW,
x grid style={darkgray176},
xlabel={\(\displaystyle t\) in \si{\second}},
xmajorgrids,
xmin=-0.995, xmax=20.895,
xtick style={color=black},
y grid style={darkgray176},
ylabel near ticks,
ylabel={\(\displaystyle n_{\mathrm{b}}\) in \si{\meter}},
ymajorgrids,
ymin=-0.05, ymax=1.05,
ytick style={color=black}
]
\addplot [very thick, steelblue31119180]
table {%
0 0
0.200000047683716 0
0.299999952316284 0.0136710405349731
0.399999976158142 0.0420644283294678
0.5 0.0828680992126465
0.600000023841858 0.133064866065979
0.700000047683716 0.189712285995483
0.799999952316284 0.250175952911377
1 0.37389862537384
1.10000002384186 0.433781147003174
1.20000004768372 0.49068820476532
1.29999995231628 0.543779611587524
1.39999997615814 0.592493295669556
1.5 0.636508703231812
1.60000002384186 0.675707817077637
1.70000004768372 0.710138440132141
1.79999995231628 0.739979386329651
1.89999997615814 0.765508413314819
2 0.787073850631714
2.09999990463257 0.805068492889404
2.20000004768372 0.81990909576416
2.29999995231628 0.832018136978149
2.40000009536743 0.841809034347534
2.5 0.849675178527832
2.59999990463257 0.855981707572937
2.70000004768372 0.861059904098511
2.79999995231628 0.865203380584717
2.90000009536743 0.868667125701904
3.09999990463257 0.874381542205811
3.59999990463257 0.887613773345947
3.79999995231628 0.893755674362183
4.09999990463257 0.904016852378845
4.90000009536743 0.932722926139832
5.09999990463257 0.939146995544434
5.30000019073486 0.945039391517639
5.5 0.950369119644165
5.69999980926514 0.955142021179199
6 0.961331605911255
6.30000019073486 0.966509461402893
6.59999990463257 0.970863461494446
7 0.975679636001587
7.40000009536743 0.979646921157837
7.90000009536743 0.98371684551239
8.39999961853027 0.987010836601257
9 0.990132331848145
9.60000038146973 0.992516040802002
10.3000001907349 0.994576215744019
11.1999998092651 0.996407628059387
12.3000001907349 0.997827410697937
13.8000001907349 0.998906850814819
15.8999996185303 0.999582052230835
19.7000007629395 0.999926567077637
19.8999996185303 0.999933004379272
};
\addlegendentry{$s_{\mathrm{d}}=40$}
\addplot [very thick, darkorange25512714]
table {%
0 0
0.200000047683716 0
0.299999952316284 0.00911509990692139
0.399999976158142 0.0280462503433228
0.5 0.0556669235229492
0.600000023841858 0.0904860496520996
0.700000047683716 0.130991339683533
0.799999952316284 0.175742626190186
0.899999976158142 0.223407506942749
1 0.272778153419495
1.20000004768372 0.372482776641846
1.29999995231628 0.421089053153992
1.39999997615814 0.46794319152832
1.5 0.512519955635071
1.60000002384186 0.554417848587036
1.70000004768372 0.593349575996399
1.79999995231628 0.629131436347961
1.89999997615814 0.661671876907349
2 0.690959453582764
2.09999990463257 0.717051386833191
2.20000004768372 0.740061283111572
2.29999995231628 0.760148406028748
2.40000009536743 0.777507424354553
2.5 0.792357921600342
2.59999990463257 0.804936647415161
2.70000004768372 0.815488338470459
2.79999995231628 0.824260711669922
2.90000009536743 0.831496596336365
3 0.837431073188782
3.09999990463257 0.842286229133606
3.20000004768372 0.846268653869629
3.29999995231628 0.849567413330078
3.5 0.854773283004761
3.79999995231628 0.861053586006165
4.09999990463257 0.867595195770264
4.30000019073486 0.872712969779968
4.5 0.87856125831604
4.69999980926514 0.88508403301239
5 0.89579701423645
5.69999980926514 0.921630501747131
6 0.931655287742615
6.19999980926514 0.937718749046326
6.40000009536743 0.943238496780396
6.59999990463257 0.948209524154663
6.90000009536743 0.954686641693115
7.19999980926514 0.960114240646362
7.5 0.964669585227966
7.80000019073486 0.968533754348755
8.19999980926514 0.972882747650146
8.69999980926514 0.977412462234497
9.30000019073486 0.981928825378418
9.89999961853027 0.985660791397095
10.5 0.988697528839111
11.1999998092651 0.991459488868713
12 0.993780374526978
12.8999996185303 0.995622873306274
14.1000003814697 0.997255802154541
15.6000003814697 0.998475790023804
17.7000007629395 0.999330639839172
19.8999996185303 0.999717116355896
};
\addlegendentry{$s_{\mathrm{d}}=60$}
\addplot [very thick, forestgreen4416044]
table {%
0 0
0.200000047683716 0
0.299999952316284 0.00683653354644775
0.399999976158142 0.0210355520248413
0.5 0.0419076681137085
0.600000023841858 0.0685298442840576
0.700000047683716 0.0999469757080078
0.799999952316284 0.135223865509033
0.899999976158142 0.173465728759766
1 0.213832139968872
1.10000002384186 0.255545735359192
1.29999995231628 0.340260624885559
1.39999997615814 0.382073163986206
1.5 0.422858715057373
1.60000002384186 0.462215185165405
1.70000004768372 0.499814510345459
1.79999995231628 0.535398960113525
1.89999997615814 0.568777561187744
2 0.599819779396057
2.09999990463257 0.628451108932495
2.20000004768372 0.654646515846252
2.29999995231628 0.678424596786499
2.40000009536743 0.699841737747192
2.5 0.718985557556152
2.59999990463257 0.735969662666321
2.70000004768372 0.750928163528442
2.79999995231628 0.764009952545166
2.90000009536743 0.775374889373779
3 0.7851881980896
3.09999990463257 0.793617963790894
3.20000004768372 0.800831079483032
3.29999995231628 0.806990146636963
3.40000009536743 0.812251567840576
3.5 0.816763401031494
3.59999990463257 0.820663690567017
3.79999995231628 0.827124357223511
4 0.832503318786621
4.5 0.845149874687195
4.69999980926514 0.850808143615723
4.90000009536743 0.857024192810059
5.19999980926514 0.86734676361084
5.5 0.878557205200195
6.19999980926514 0.90532112121582
6.5 0.915919065475464
6.80000019073486 0.925535440444946
7.09999990463257 0.934037923812866
7.40000009536743 0.941413879394531
7.69999980926514 0.947739601135254
8 0.953144550323486
8.30000019073486 0.957781553268433
8.69999980926514 0.963028192520142
9.10000038146973 0.967482089996338
9.60000038146973 0.972266793251038
10.1999998092651 0.977180004119873
10.8000001907349 0.981358289718628
11.3999996185303 0.984862565994263
12.1000003814697 0.98816180229187
12.8000001907349 0.990726828575134
13.6999998092651 0.993189811706543
14.6999998092651 0.995149612426758
16 0.996886730194092
17.6000003814697 0.998203039169312
19.7000007629395 0.999124526977539
19.8999996185303 0.999182462692261
};
\addlegendentry{$s_{\mathrm{d}}=80$}
\addplot [very thick, crimson2143940]
table {%
0 0
0.200000047683716 0
0.299999952316284 0.00546932220458984
0.399999976158142 0.0168287754058838
0.5 0.0336015224456787
0.600000023841858 0.0551429986953735
0.700000047683716 0.0807777643203735
0.799999952316284 0.109833598136902
0.899999976158142 0.141655802726746
1 0.17561674118042
1.10000002384186 0.211123466491699
1.20000004768372 0.247624039649963
1.5 0.358262777328491
1.60000002384186 0.394161224365234
1.70000004768372 0.42901599407196
1.79999995231628 0.462570786476135
1.89999997615814 0.494617342948914
2 0.524993300437927
2.09999990463257 0.553579330444336
2.20000004768372 0.580296039581299
2.29999995231628 0.60509991645813
2.40000009536743 0.627980470657349
2.5 0.648956298828125
2.59999990463257 0.668070554733276
2.70000004768372 0.685387969017029
2.79999995231628 0.700991272926331
2.90000009536743 0.714977025985718
3 0.727452754974365
3.09999990463257 0.738534212112427
3.20000004768372 0.748341679573059
3.29999995231628 0.756998300552368
3.40000009536743 0.76462733745575
3.5 0.771350145339966
3.59999990463257 0.777284622192383
3.70000004768372 0.782543540000916
3.79999995231628 0.787233829498291
4 0.795299530029297
4.19999980926514 0.80218243598938
4.59999990463257 0.814525127410889
4.90000009536743 0.823910593986511
5.19999980926514 0.834061503410339
5.5 0.845076560974121
5.90000009536743 0.860741853713989
6.5 0.884536504745483
6.80000019073486 0.895755767822266
7.09999990463257 0.906151533126831
7.40000009536743 0.915570497512817
7.69999980926514 0.923957347869873
8 0.93133556842804
8.30000019073486 0.937785863876343
8.60000038146973 0.943422079086304
8.89999961853027 0.948370575904846
9.30000019073486 0.95410943031311
9.69999980926514 0.95909309387207
10.1999998092651 0.964539170265198
10.8000001907349 0.970209121704102
11.3999996185303 0.975101351737976
12 0.97928512096405
12.6999998092651 0.983332276344299
13.3999996185303 0.986581444740295
14.1999998092651 0.989494800567627
15.1000003814697 0.991997838020325
16.2000007629395 0.994259357452393
17.5 0.99613356590271
19.1000003814697 0.997626066207886
19.8999996185303 0.998138666152954
};
\addlegendentry{$s_{\mathrm{d}}=100$}
\addplot [very thick, mediumpurple148103189]
table {%
0 0
0.200000047683716 0
0.299999952316284 0.00455784797668457
0.399999976158142 0.0140241384506226
0.5 0.0280431509017944
0.600000023841858 0.0461297035217285
0.700000047683716 0.067771315574646
0.799999952316284 0.0924522876739502
0.899999976158142 0.119664430618286
1 0.148914575576782
1.10000002384186 0.179730653762817
1.20000004768372 0.211666464805603
1.39999997615814 0.277266621589661
1.60000002384186 0.342795968055725
1.70000004768372 0.374779462814331
1.79999995231628 0.405914068222046
1.89999997615814 0.435999274253845
2 0.464869976043701
2.09999990463257 0.492394924163818
2.20000004768372 0.51847505569458
2.29999995231628 0.543040633201599
2.40000009536743 0.566049814224243
2.5 0.587485551834106
2.59999990463257 0.607352733612061
2.70000004768372 0.62567663192749
2.79999995231628 0.642498970031738
2.90000009536743 0.657876372337341
3 0.671877145767212
3.09999990463257 0.684579133987427
3.20000004768372 0.696067810058594
3.29999995231628 0.706433534622192
3.40000009536743 0.715770125389099
3.5 0.724172830581665
3.59999990463257 0.731737017631531
3.70000004768372 0.738556385040283
3.79999995231628 0.744722008705139
3.90000009536743 0.750321507453918
4 0.755437731742859
4.19999980926514 0.764526128768921
4.40000009536743 0.772539377212524
4.69999980926514 0.783505082130432
5.30000019073486 0.805106639862061
5.59999990463257 0.816628813743591
6 0.832789182662964
6.80000019073486 0.86560583114624
7.09999990463257 0.877183675765991
7.40000009536743 0.888002395629883
7.69999980926514 0.897932410240173
8 0.90691876411438
8.30000019073486 0.914970755577087
8.60000038146973 0.922145009040833
8.89999961853027 0.928529024124146
9.19999980926514 0.93422532081604
9.60000038146973 0.940929651260376
10 0.946818828582764
10.5 0.953301429748535
11 0.959024786949158
11.6000003814697 0.965076088905334
12.1999998092651 0.970339298248291
12.8000001907349 0.974872589111328
13.5 0.97931432723999
14.1999998092651 0.982955455780029
15 0.986310005187988
15.8999996185303 0.989281535148621
16.8999996185303 0.991832256317139
18.1000003814697 0.99411416053772
19.5 0.995988130569458
19.8999996185303 0.996403932571411
};
\addlegendentry{$s_{\mathrm{d}}=120$}
\addplot [very thick, sienna1408675]
table {%
0 0
0.200000047683716 0
0.299999952316284 0.0039067268371582
0.399999976158142 0.0120207071304321
0.5 0.0240625143051147
0.600000023841858 0.039648175239563
0.700000047683716 0.058369517326355
0.799999952316284 0.07981276512146
0.899999976158142 0.103566884994507
1 0.129229426383972
1.10000002384186 0.156411647796631
1.20000004768372 0.184742569923401
1.29999995231628 0.213872909545898
1.70000004768372 0.332277297973633
1.79999995231628 0.361062407493591
1.89999997615814 0.389106273651123
2 0.416251182556152
2.09999990463257 0.442366600036621
2.20000004768372 0.467348098754883
2.29999995231628 0.491116285324097
2.40000009536743 0.513614416122437
2.5 0.534807443618774
2.59999990463257 0.554679751396179
2.70000004768372 0.573233127593994
2.79999995231628 0.590484976768494
2.90000009536743 0.60646641254425
3 0.621220111846924
3.09999990463257 0.634798288345337
3.20000004768372 0.647261738777161
3.29999995231628 0.658676862716675
3.40000009536743 0.669114828109741
3.5 0.678650140762329
3.59999990463257 0.687358856201172
3.70000004768372 0.695317625999451
3.79999995231628 0.702602386474609
3.90000009536743 0.709287643432617
4 0.715445280075073
4.09999990463257 0.721144437789917
4.30000019073486 0.731423735618591
4.5 0.74059534072876
4.80000019073486 0.753117799758911
5.80000019073486 0.793288707733154
6.40000009536743 0.818816661834717
6.90000009536743 0.84001624584198
7.30000019073486 0.85614013671875
7.59999990463257 0.867446660995483
7.90000009536743 0.877944111824036
8.19999980926514 0.887574911117554
8.5 0.896334886550903
8.80000019073486 0.904260993003845
9.10000038146973 0.91141939163208
9.39999961853027 0.917892217636108
9.80000019073486 0.925608158111572
10.1999998092651 0.932458877563477
10.6000003814697 0.938611745834351
11.1000003814697 0.945513248443604
11.6000003814697 0.95168936252594
12.1999998092651 0.958276510238647
12.8000001907349 0.964047908782959
13.3999996185303 0.969064593315125
14 0.973390102386475
14.6999998092651 0.97766375541687
15.5 0.981688737869263
16.3999996185303 0.985338449478149
17.3999996185303 0.988545536994934
18.5 0.991276502609253
19.7999992370605 0.993681907653809
19.8999996185303 0.993836760520935
};
\addlegendentry{$s_{\mathrm{d}}=140$}
\addplot [very thick, black, dashed]
table {%
0 1
19.8999996185303 1
};
\addlegendentry{$n_{\mathrm{o}}(t)$}
\end{axis}

\end{tikzpicture}

%% file: chapters/methodology.tex
\section{Methodology}
\label{sec:methodology}
The conventional sampling-based approach for trajectory planning of the overtaking vehicle is described in Section~\ref{subec:conventional_approach}. Our proposed \ac{RL}-based approach with the \ac{SL} follows in Section~\ref{subec:rl_based_approach} and \ref{subec:safety_layer}.

\subsection{Conventional Trajectory Planning Approach}
\label{subec:conventional_approach}
The conventional trajectory planning method is a sampling-based approach, as applied in \cite{Jung2023, Raji2024, Oegretmen2024b}. A set of trajectory candidates is generated, the individual trajectories are checked for feasibility, and then the optimal valid trajectory with respect to a cost function is selected. The trajectory candidates are generated by constructing a set of longitudinal $\mathcal{T}_{\mathrm{long}}$ and lateral curves $\mathcal{T}_{\mathrm{lat}}$ in curvilinear coordinates and combining each element of the two sets $\mathcal{T}_{\mathrm{long}} \times \mathcal{T}_{\mathrm{lat}}$. The sets are constructed by sampling end states and then connecting the vehicle state $\left[s_{\mathrm{o}}, \dot{s}_{\mathrm{o}}, \ddot{s}_{\mathrm{o}}, n_{\mathrm{o}}, \dot{n}_{\mathrm{o}}, \ddot{n}_{\mathrm{o}}\right]$ to them using jerk-minimal curves with a fixed temporal horizon $T$. 

For the lateral curves $n_i(t)$, we vary the end state $\mathcal{N}_{\mathrm{e}} = \left[n_{\mathrm{e},i}, \dot{n}_{\mathrm{e}}, \ddot{n}_{\mathrm{e}}\right]$ by sampling $N_{\mathrm{n}}$ equidistantly distributed positions within the track bounds $n_{\mathrm{e},i} \in \left[ n_{\mathrm{r}} + \sfrac{d_{\mathrm{w}}}{2}, \, n_{\mathrm{l}} - \sfrac{d_{\mathrm{w}}}{2} \right]$, considering the vehicle width $d_{\mathrm{w}}$. Further, the end velocity $\dot{n}_{\mathrm{e}}$ and acceleration $\ddot{n}_{\mathrm{e}}$ are set to \num{0} to escape the curse of dimensionality. Quintic polynomials, which are proven to be jerk-minimal \cite{Werling2010}, are used to connect the vehicle state to the sampled end states $\mathcal{N}_{\mathrm{e}}$. Similarly, for the longitudinal curves $s_j(t)$, we vary the end state $\mathcal{S}_{\mathrm{e}} = \left[\dot{s}_{\mathrm{e},j}, \ddot{s}_{\mathrm{e}}\right]$ by sampling $N_{\mathrm{\dot{s}}}$ equidistantly distributed velocities in the interval $\dot{s}_{\mathrm{e},j} \in \left[ 0, v_{\mathrm{\max}}\right]$. Again, $\ddot{s}_{\mathrm{e}}=0$ is chosen. Since the end position is not specified, the end state lies on a manifold, for which quartic polynomials are jerk-minimal \cite{Werling2010}.

Each combination of lateral and longitudinal curve $\left( n_i(t), \, s_j(t) \right)$ is transformed into Cartesian coordinates, and the resulting trajectory is checked for feasibility. For this, we use the same three feasibility checks as in \cite{Oegretmen2024b}. First, it is checked whether the trajectory leads to a collision with the track boundaries. Secondly, as a kinematic constraint, it is checked whether the vehicle's minimal turning radius $r_{\min}$ is not violated. Finally, as a dynamic constraint, it is checked whether the combined lateral and longitudinal acceleration lies within gg-diagrams calculated offline. In contrast to \cite{Oegretmen2024b}, in which the three-dimensional geometry of the race track impacts the gg-diagrams, here only a dependency on the velocity $v_{\mathrm{o}}(t)$ is present due to the two-dimensional flat track.

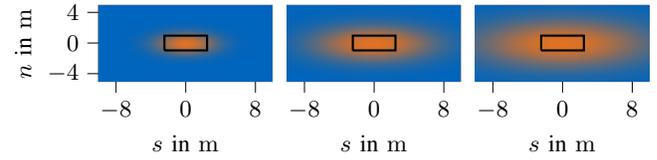
\begin{figure}[t]
	\centering
	\small
	\input{figures/predictioncost.tex}
	\caption{Elliptical prediction cost shape for different parameterizations. The costs range from $d_{\mathrm{pr}}=\num{0}$ (blue) to $d_{\mathrm{pr}}=\num{1}$ (orange). The actual vehicle geometry is depicted in black. From left to right: small ellipse $(p_{\mathrm{s}}=\num{0.08}, p_{\mathrm{n}}=\num{0.5})$, medium ellipse $(p_{\mathrm{s}}=\num{0.02}, p_{\mathrm{n}}=\num{0.18})$, and large ellipse $(p_{\mathrm{s}}=\num{0.01}, p_{\mathrm{n}}=\num{0.1})$.}
	\label{fig:prediction_cost}
\end{figure}

For the design of the cost function, we follow \cite{Oegretmen2024b} and calculate the scalar cost $C$ of a valid trajectory as
\begin{equation}
	C = \int_{0}^{T} w_{\mathrm{n}} \cdot n_{\mathrm{o}}^2(t) + w_{\mathrm{v}} \cdot (v_{\max} - v_{\mathrm{o}}(t))^2 + w_{\mathrm{pr}} \cdot d_{\mathrm{pr}}(t) \,\, \mathrm{d} t
	\label{eq:cost_function}
\end{equation}
with the weighting parameters $w_{\mathrm{n}}$, $w_{\mathrm{v}}$, and $w_{\mathrm{pr}}$. The first term penalizes lateral deviations from the reference line, and the second term the deviations from the maximum vehicle velocity $v_{\max}$. The last term penalizes trajectories that are close to the prediction $\left( s_{\mathrm{b,pr}}(t), n_{\mathrm{b,pr}}(t) \right)$ of the blocking vehicle, aiming to avoid collisions. Using the formulation
\begin{equation}
	d_{\mathrm{pr}}(t) = e^{- p_{\mathrm{s}} \left[ s_{\mathrm{b,pr}}(t) - s_{\mathrm{o}}(t) \right]^2 - p_{\mathrm{n}} \left[ n_{\mathrm{b,pr}}(t) - n_{\mathrm{o}}(t) \right]^2 },
	\label{eq:prediction_cost}
\end{equation}
an elliptical cost shape is spanned around the predicted position for each time step. The length and width can be adjusted by the parameters $p_{\mathrm{s}}$ and $p_{\mathrm{n}}$, as shown in Fig.~\ref{fig:prediction_cost}.

\setlength{\figH}{0.3\columnwidth}
\setlength{\figW}{0.45\columnwidth}
\setlength{\horiDis}{0.2cm}

According to \eqref{eq:prediction_cost}, the prediction significantly influences the decision on the optimal trajectory. Since the acceleration $a_{\mathrm{b}}$ of the blocking vehicle is set to \num{0} as described in Section~\ref{sec:blocking_scenario}, a constant velocity model is used for the longitudinal motion prediction. For lateral movement, however, a reliable prediction is not possible, as the opponent's behavior depends on the decision of the overtaking vehicle. As these are commonly used prediction approaches, we investigate both the assumption of a \acf{CH} and \acf{CLP}, with clipping of $n_{\mathrm{b,pr}}(t)$, preventing the track boundaries from being surpassed.

\subsection{RL-based Trajectory Planning Approach}
\label{subec:rl_based_approach}

The \ac{RL}-based planning approach uses the same jerk-optimal trajectory generation as the conventional approach in Section~\ref{subec:conventional_approach}. However, instead of generating a set of trajectories by sampling end states, the end state is explicitly selected by the \ac{RL} agent, similar to \cite{Mirchevska2023}. This generates only a single trajectory, avoiding the need to define a cost function and the associated prediction of the blocking vehicle. A feasibility check of the planned trajectory is carried out using the same checks as the conventional approach. The corresponding \ac{MDP} formulation and the utilized training procedure are presented below.

\subsubsection{State Space}
The state space of the \ac{MDP} contains the vehicle state of the overtaking vehicle in curvilinear coordinates $\left[s_{\mathrm{o}}, \dot{s}_{\mathrm{o}}, \ddot{s}_{\mathrm{o}}, n_{\mathrm{o}}, \dot{n}_{\mathrm{o}}, \ddot{n}_{\mathrm{o}}\right]$ and its relative orientation $\chi_\mathrm{o}$ to the reference line. It additionally includes the relative values to the blocking vehicle $\left[s_{\mathrm{o}}-s_{\mathrm{b}}, \dot{s}_{\mathrm{o}}-\dot{s}_{\mathrm{b}}, n_{\mathrm{o}}-n_{\mathrm{b}}, \dot{n}_{\mathrm{o}}-\dot{n}_{\mathrm{b}}, \chi_\mathrm{o}-\chi_\mathrm{b}\right]$. All variables are normalized to lie between \num{-1} and \num{1} by dividing them by the maximum possible value to accelerate the training process.

\subsubsection{Action Space}
The action space represents the end state of the trajectory in curvilinear coordinates. Although it would be conceivable to consider each end condition, we omit $\ddot{s}_{\mathrm{e}}$ in the action space and choose $\ddot{s}_{\mathrm{e}}=0$ instead. Otherwise, negative end accelerations would be selected, which allow higher accelerations at the beginning of the trajectory but do not correspond to a desired realistic trajectory. In addition, the end time $T$ is also not included to ensure a constant planning horizon. The four-dimensional action space is thus chosen as $\left[n_{\mathrm{e}}, \dot{n}_{\mathrm{e}}, \ddot{n}_{\mathrm{e}}, \dot{s}_{\mathrm{e}}\right]$ and normalized like the state space. Note that in the conventional approach, $\dot{n}_{\mathrm{e}}=\ddot{n}_{\mathrm{e}}=0$ is set for reasons of computational effort. Thus, a larger variety of trajectories can be generated with the \ac{RL}-based approach.

\subsubsection{Reward Function}
The agent receives in each time step $t$ a total scalar reward $r_t$, which has been designed empirically. It comprises sparse terminal rewards received at the end of an episode and a dense reward $r_{\mathrm{d}}$ during an episode that guides the agent toward the desired behavior:
\begin{equation}
	r_t = \begin{cases}
		-1, & \text{if episode ends unsuccessful} \\
		10, & \text{if episode ends successful} \\
		r_{\mathrm{d}}, & \text{otherwise}.
	\end{cases}
\end{equation}
The dense reward $r_{\mathrm{d}}$ is intended to guide the agent to successful overtakes and consists of the following two terms:
\begin{equation}
	\begin{split}
		r_{\mathrm{d}} = &\frac{1}{2} \left(|\Delta n| - k_{\mathrm{scl}} d_{\mathrm{w}}\right) \cdot \mathds{1}\left(s_{\mathrm{o}} \in \left[s_{\mathrm{b}} - d_{\mathrm{l}}, s_{\mathrm{b}} + d_{\mathrm{l}}\right]\right) \\
		&+ \left(\Delta \dot{s} - \Delta \dot{s}_{\mathrm{max}}\right) \cdot \mathds{1}\left(\Delta \dot{s} > \Delta \dot{s}_{\mathrm{max}}\right).
	\end{split}
	\label{eq:dense_reward}
\end{equation}
In this formulation, we use an indicator that specifies whether a specific condition is fulfilled:
\begin{equation}
	\mathds{1}\left(\cdot\right) = \begin{cases}
		1, & \text{if $(\cdot)$ is True} \\
		0, & \text{if $(\cdot)$ is False}.
	\end{cases}
\end{equation}
The first term in \eqref{eq:dense_reward} rewards large lateral distances $|\Delta n| = |n_{\mathrm{o}} - n_{\mathrm{b}}|$ between the overtaking and blocking vehicle and is only active if the two vehicles of length $d_{\mathrm{l}}$ overlap in the longitudinal direction. Negative rewards result from distances that also lead to a lateral overlap. The vehicle width $d_{\mathrm{w}}$ is scaled with the scaling factor $k_{\mathrm{scl}} \in \left[0,1\right]$, which is utilized in the training procedure in Section~\ref{subsubsec:training}. The second term rewards high longitudinal relative velocities $\Delta \dot{s} = \dot{s}_{\mathrm{o}} - \dot{s}_{\mathrm{b}}$ and is only active if the current relative velocity exceeds the maximum value $\Delta \dot{s}_{\mathrm{max}}$ up to this time step.

\subsubsection{Training Algorithm and Network Architecture}
For training, we utilize the widely used \ac{PPO} algorithm described in Section~\ref{sec:rl_preliminaries} consisting of an actor (policy) and a critic (value function). For both, we use fully connected neural networks as function approximators. The policy network maps from the state space (input layer dimension \num{12}) to the action space (output layer dimension \num{4}). The value function network also maps from the state space (input layer dimension \num{12}) to the value of the state (output layer dimension \num{1}). Both networks have two hidden layers and \num{256} neurons per layer using the TanH activation function.

\subsubsection{Training Procedure}
\label{subsubsec:training}
We use a multi-stage curriculum learning procedure for faster and improved training in which the task's difficulty increases with each stage. In the first stage, we train the agent to generate feasible trajectories without accounting for the blocking vehicle. Accordingly, no collision checks are performed, and the term for lateral distances in the reward function \eqref{eq:dense_reward} is omitted.

In the subsequent stages, the blocking vehicle is considered, and thus, a strategy is trained to perform successful overtaking maneuvers despite the interactive blocking behavior. In these stages, the vehicle width and length are scaled by a factor $k_{\mathrm{scl}} \in \left\{0.2, 0.4, \dots, 1.0\right\}$ within the collision checks. With each stage, the scaling factor and, thus, the difficulty level is gradually increased until the actual vehicle geometry is used in the final sixth stage. The scaling factor $k_{\mathrm{scl}}$ is correspondingly incorporated into the lateral distance term in the reward function as shown in \eqref{eq:dense_reward}.

\subsection{Safety Layer}
\label{subec:safety_layer}
A fundamental problem with \ac{RL}-based approaches is that the feasibility of the selected trajectory is not guaranteed. Especially in scenarios not included in the training, the probability of infeasibility increases. Instead of terminating the scenario unsuccessfully in such a case as in \cite{Mirchevska2023}, we introduce an \ac{SL} intended to increase robustness and, thus, the success rate.  

For this, we generate the same set of trajectories used in Section~\ref{subec:conventional_approach} for the conventional approach when the trajectory $\left(s_{\mathrm{\acs{RL}}}(t), n_{\mathrm{\acs{RL}}}(t)\right)$ selected by the \ac{RL}-based approach is infeasible. All trajectories within the set are checked for feasibility so that only valid trajectories remain. In contrast to the conventional approach, cost function \eqref{eq:cost_function} is not used here to select the optimal trajectory. Instead, the trajectory most similar to the infeasible one chosen by the \ac{RL} agent is selected. This is done using the cost function
\begin{equation}
	C_\mathrm{\acs{SL}} = \int_{0}^{T} \left(s_{\mathrm{\acs{RL}}}(t) - s_{\mathrm{o}}(t)\right)^2 + \left(n_{\mathrm{\acs{RL}}}(t) - n_{\mathrm{o}}(t)\right)^2 \,\, \mathrm{d} t.
	\label{eq:cost_function_sl}
\end{equation}

%% file: figures/predictioncost.tex
\begin{tikzpicture}

\definecolor{darkgray176}{RGB}{176,176,176}

\begin{groupplot}[group style={group size=3 by 1, horizontal sep=\horiDis}]
\nextgroupplot[
height=\figH,
tick align=outside,
tick pos=left,
width=\figW,
x grid style={darkgray176},
xlabel={\(\displaystyle s\) in \si{\meter}},
xmin=-10, xmax=9.89999999999993,
xtick style={color=black},
xtick={-8, 0, 8},
y grid style={darkgray176},
ylabel near ticks,
ylabel={\(\displaystyle n\) in \si{\meter}},
ymin=-5, ymax=4.89999999999996,
ytick style={color=black},
ytick={-4, 0, 4}
]
\addplot graphics [includegraphics cmd=\pgfimage,xmin=-10, xmax=9.89999999999993, ymin=-5, ymax=4.89999999999996] {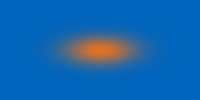};
\path [draw=black, thick]
(axis cs:-2.45,-0.965)
--(axis cs:-2.45,0.965)
--(axis cs:2.45,0.965)
--(axis cs:2.45,-0.965)
--(axis cs:-2.45,-0.965);

\nextgroupplot[
height=\figH,
scaled y ticks=manual:{}{\pgfmathparse{#1}},
tick align=outside,
width=\figW,
x grid style={darkgray176},
xlabel={\(\displaystyle s\) in \si{\meter}},
xmin=-10, xmax=9.89999999999993,
xtick pos=left,
xtick style={color=black},
xtick={-8, 0, 8},
y grid style={darkgray176},
ylabel near ticks,
ymajorticks=false,
ymin=-5, ymax=4.89999999999996,
ytick style={color=black},
ytick={-4, 0, 4},
yticklabels={}
]
\addplot graphics [includegraphics cmd=\pgfimage,xmin=-10, xmax=9.89999999999993, ymin=-5, ymax=4.89999999999996] {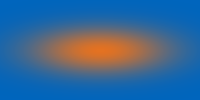};
\path [draw=black, thick]
(axis cs:-2.45,-0.965)
--(axis cs:-2.45,0.965)
--(axis cs:2.45,0.965)
--(axis cs:2.45,-0.965)
--(axis cs:-2.45,-0.965);

\nextgroupplot[
height=\figH,
scaled y ticks=manual:{}{\pgfmathparse{#1}},
tick align=outside,
width=\figW,
x grid style={darkgray176},
xlabel={\(\displaystyle s\) in \si{\meter}},
xmin=-10, xmax=9.89999999999993,
xtick pos=left,
xtick style={color=black},
xtick={-8, 0, 8},
y grid style={darkgray176},
ylabel near ticks,
ymajorticks=false,
ymin=-5, ymax=4.89999999999996,
ytick style={color=black},
ytick={-4, 0, 4},
yticklabels={}
]
\addplot graphics [includegraphics cmd=\pgfimage,xmin=-10, xmax=9.89999999999993, ymin=-5, ymax=4.89999999999996] {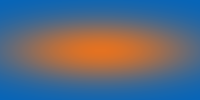};
\path [draw=black, thick]
(axis cs:-2.45,-0.965)
--(axis cs:-2.45,0.965)
--(axis cs:2.45,0.965)
--(axis cs:2.45,-0.965)
--(axis cs:-2.45,-0.965);
\end{groupplot}

\end{tikzpicture}

%% file: chapters/results_discussion.tex
\section{Results}
\label{sec:results_discussion}
We evaluate the conventional and the \ac{RL}-based planning approach from Section~\ref{sec:methodology}. We start in Section~\ref{subsec:computational_details} with computational details and explain the evaluation method used in Section~\ref{subsec:evaluation_method}. The actual evaluation results for the two approaches follow in Sections~\ref{subsec:result_conventional} and \ref{subsec:result_rl_based}. Finally, the influence of the \ac{SL} is examined in Section~\ref{subsec:result_safety:layer}.

\subsection{Computational Details}
\label{subsec:computational_details}
We use gymnasium\footnote{\url{https://gymnasium.farama.org}} for the implementation of the environment and the \ac{PPO} implementation from Stable-Baselines3 \cite{Raffin2021}. The generated trajectories are discretized into $N$ equidistantly distributed points in time, and the cost functions \eqref{eq:cost_function} and \eqref{eq:cost_function_sl} are approximated according to the rectangle rule. 

For the forward propagation of the blocking vehicle, the system \eqref{eq:system_dynamics} is discretized with a step size of $\Delta t$ using the forward Euler method. The overtaking vehicle is propagated forward using the same time step $\Delta t$ with the assumption of perfect tracking of the planned trajectory. All experiment parameters used are listed in Table~\ref{tab:parameters}.

\input{tables/experiment_parameters.tex}

\subsection{Evaluation Method}
\label{subsec:evaluation_method}
We evaluate both approaches in the blocking scenario introduced in Section~\ref{sec:blocking_scenario}. To evaluate the generalization to different situations, we vary the initialization and the aggressiveness of the blocking vehicle. The second column in Table~\ref{tab:evaluation_training_parameters} shows the variation of the corresponding parameters. As stated, the initial position of the blocking vehicle $(s_\mathrm{b, init}, n_\mathrm{b, init})$ is varied for all six blocking behaviors shown in Fig.~\ref{fig:step_responses}, resulting in $\num{1722}$ configurations. In the following sections, we specify the success rate for each blocking behavior, i.e., for each value of $s_{\mathrm{d}}$. This success rate, therefore, indicates the percentage of different initializations for which a successful overtaking maneuver could be executed.

\input{tables/evaluation_parameters.tex}

\subsection{Conventional Approach}
\label{subsec:result_conventional}
The performance of the conventional approach depends significantly on the prediction method used and the selection of the parameters within the cost function in \eqref{eq:cost_function} and \eqref{eq:prediction_cost}. Since a complete factorial search for the optimal cost function weight set is not possible due to the number of degrees of freedom, we evaluate six variants of the conventional approach listed in Table~\ref{tab:cost_parameters}. Here, the three differently sized ellipses from Fig.~\ref{fig:prediction_cost} are used, specifying the parameters $p_{\mathrm{s}}$ and $p_{\mathrm{n}}$. Since only the relative ratio of the three remaining cost weighting parameters is relevant, we choose $w_{\mathrm{pr}}$ fixed for all variants. Finally, $w_{\mathrm{n}}$ and $w_{\mathrm{v}}$ are determined by a full factorial search in the value range of $[\num{0}, \num{1}]$ with a step size of \num{0.04}. The two options for a given ellipse shape result from a prediction based on \acf{CH} or \acf{CLP}.

\input{tables/cost_parameters.tex}

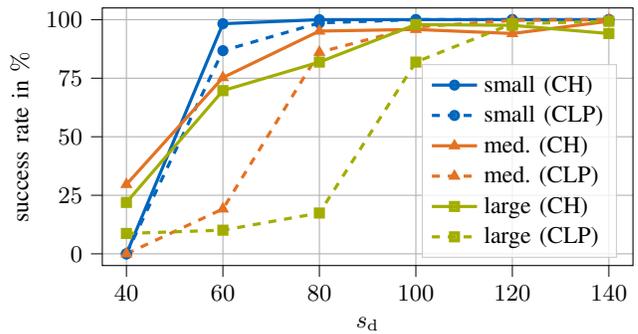
\begin{figure}[t]
	\centering
	\small
	\input{figures/conventional_evaluation.tex}
	\caption{Evaluation of the conventional approach with the parameterizations listed in Table~\ref{tab:cost_parameters} for different blocking behaviors specified by $s_{\mathrm{d}}$. The terms small, medium (med.), and large correspond to the different ellipse sizes.}
	\label{fig:conventional_evaluation}
\end{figure}

Fig.~\ref{fig:conventional_evaluation} shows the success rates of the respective variants for the different blocking behaviors. Smaller values of $s_{\mathrm{d}}$ correspond to more aggressive blocking behavior and, thus, an increased interaction. Due to the associated higher discrepancy between actual blocking behavior and prediction, the success rate generally decreases with smaller values of $s_{\mathrm{d}}$. In addition, the \ac{CH} prediction is more suitable than the \ac{CLP} prediction for almost all blocking behaviors. The highest success rates of nearly \SI{100}{\percent} for most behaviors result for the small ellipse with \ac{CH} prediction. Since this cost parameterization causes low prediction costs for a large area of the race track, most overtaking maneuvers are initiated immediately, and conservative behavior is avoided. For $s_{\mathrm{d}} = \num{40}$, however, no successful overtaking maneuver can be executed since the blocking vehicle reacts at a rate that makes collision avoidance impossible with this strategy. Instead of immediately initiating an overtaking maneuver, the blocking vehicle would have to be pulled to one side of the race track to allow for an overtake on the other side with more space. However, this targeted exploitation of the interaction is not possible with the conventional approach.

\setlength{\figH}{0.58\columnwidth}
\setlength{\figW}{1.0\columnwidth}

\subsection{RL-based Approach}
\label{subsec:result_rl_based}
During training, we use only a subset of the values used in the evaluation for the parameter $s_{\mathrm{d}}$ to investigate the generalization behavior of the \ac{RL}-based approach to different aggressiveness levels of blocking. As listed in the third and fourth columns of Table~\ref{tab:evaluation_training_parameters}, we specifically distinguish two \ac{RL} agents that differ in the values of $s_{\mathrm{d}}$ used during the training process. While only one value for $s_{\mathrm{d}}$ and thus only one blocking behavior is considered in training \#1, three of the six different blocking behaviors are included in training \#2. The starting position of the blocking vehicle $(s_\mathrm{b, init}, n_\mathrm{b, init})$, in contrast, is randomly sampled at the intervals specified in the table. The interval limits are selected as the extreme values examined in the evaluation, meaning there is no discrepancy between training and evaluation.

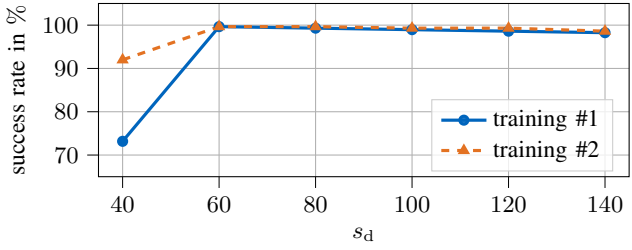
\begin{figure}[t]
	\centering
	\small
	\input{figures/rl_based_evaluation.tex}
	\caption{Evaluation of the \ac{RL}-based approach with the training parameterizations listed in Table~\ref{tab:evaluation_training_parameters} for different blocking behaviors specified by $s_{\mathrm{d}}$.}
	\label{fig:rl_based_evaluation}
\end{figure}

Fig.~\ref{fig:rl_based_evaluation} shows the success rates for the two \ac{RL} agents after training. The trajectories generated by both agents are always feasible in the examined scenarios, meaning an unsuccessful episode only occurs due to a collision. While there are only slight differences for most $s_{\mathrm{d}}$ values, more significant differences occur at $s_{\mathrm{d}}=40$, i.e., at the highest interaction level. While the success rate for agent \#2 remains high with values over \SI{90}{\percent}, it is significantly lower for agent \#1, where only one value of $s_{\mathrm{d}}$ is used in the training. This demonstrates the importance of various opponent behaviors during training. For the remaining less aggressive blocking behaviors not included in the training, successful generalization is achieved with success rates close to \SI{100}{\percent}.

\setlength{\figH}{0.45\columnwidth}
\setlength{\figW}{1.0\columnwidth}

\begin{figure}[t]
	\centering
	\small
	\input{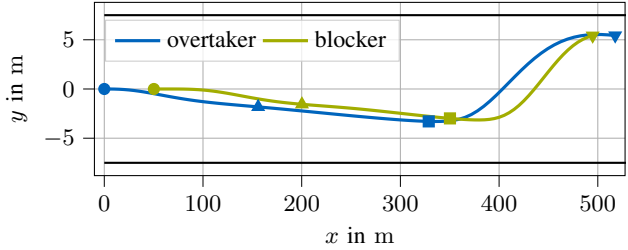}
	\caption{Overtake using the \ac{RL}-based approach with training parameterization \#2 against a blocking opponent with $s_{\mathrm{d}}=40$, $s_\mathrm{b, init} = \SI{50}{\meter}$ and $n_\mathrm{b, init} = \SI{0}{\meter}$. The track bounds are indicated in black, and markers of the same shape represent the same point in time.}
	\label{fig:rl_vs_opp}
\end{figure}

Compared to the conventional approach, the success rates are higher, especially for smaller values of $s_{\mathrm{d}}$. While the conventional approach reaches a maximum success rate of \SI{30}{\percent} for $s_{\mathrm{d}}=40$, the \ac{RL}-based approach achieves up to \SI{92}{\percent}. This is realized by exploiting the interaction between the two vehicles, as shown in Fig.~\ref{fig:rl_vs_opp}. The overtaking vehicle initially swerves to one side of the track, causing the blocking vehicle to move accordingly laterally. The space freed up on the other side is then exploited for a successful overtaking maneuver. In contrast, such a maneuver is not possible with the conventional approach, demonstrating the drawback of a sequential architecture consisting of prediction and planning. Furthermore, the \ac{RL}-based approach has a mean computing time per planning cycle of \SI{1.5}{\milli\second}, which is \num{17} times shorter than the conventional approach requiring \SI{26}{\milli\second}.

\setlength{\figH}{0.45\columnwidth}
\setlength{\figW}{1.0\columnwidth}

\subsection{Safety Layer}
\label{subsec:result_safety:layer}
The \ac{RL} agent learns to generate feasible trajectories during training. However, if a scenario deviates more strongly from the scenarios contained in the training, previously unseen states occur, which can lead to infeasibility. To prevent the episode from ending unsuccessfully in such a case, the \ac{SL} presented in Section~\ref{subec:safety_layer} generates a valid trajectory close to the infeasible one.

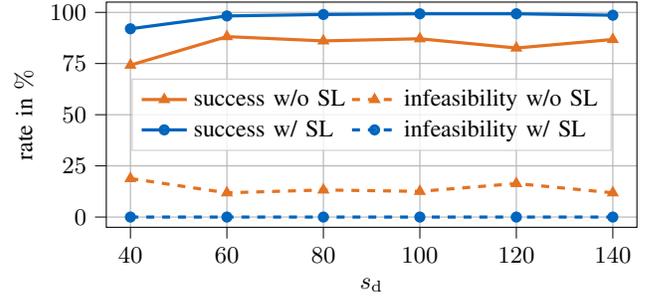
\begin{figure}[t]
	\centering
	\small
	\input{figures/safety_layer_evaluation.tex}
	\caption{Evaluation of the \ac{RL}-based approach without (w/o) and with (w/) the use of the \ac{SL} for different blocking behaviors specified by $s_{\mathrm{d}}$. The training parameterization \#2 listed in Table~\ref{tab:evaluation_training_parameters} is used with an additive gaussian noise for the opponent velocity estimation of mean $\mu=\SI{0}{\meter\per\second}$ and standard deviation $\sigma=\SI{0.7}{\meter\per\second}$.}
	\label{fig:safety_layer_evaluation}
\end{figure}

To investigate the effect of the \ac{SL}, we rerun the evaluation for the \ac{RL}-based approach but include additive Gaussian noise to the estimation of the opponent velocity $\dot{s}_{\mathrm{b}}$, which corresponds to sensor noise in a real-world application. The noise causes states not included during training, resulting in infeasible trajectories. Fig.~\ref{fig:safety_layer_evaluation} shows the corresponding success and infeasibility rates both with and without using the \ac{SL}. Here, the infeasibility rate indicates the percentage of episodes aborted due to an infeasible trajectory. Without the \ac{SL}, an infeasibility rate of around \SI{15}{\percent} results and a correspondingly low success rate. However, the infeasibility issues are entirely counteracted by the \ac{SL}, achieving success rates close to the evaluation without the additional noise. The remaining unsuccessful episodes result from collisions with the opposing vehicle. However, these cases cannot be handled like infeasibility in the \ac{SL}, as a collision check of a trajectory requires a prediction of the opposing vehicle, which is avoided in the \ac{RL}-based approach.

\setlength{\figH}{0.5\columnwidth}
\setlength{\figW}{1.0\columnwidth}


%% file: tables/experiment_parameters.tex
\begin{table}[t]
	\caption{Experiment Parameters}
	\begin{center}
		\begin{tabular}{c c}
			\toprule 
			Parameter & Value \\
			\midrule
			$T$ & \SI{2.5}{\second} \\
			$N_{\mathrm{\dot{s}}}$ & \num{40} \\
			$N_{\mathrm{n}}$ & \num{20} \\
			$N$ & \num{51} \\
			$\Delta t$ & \SI{0.1}{\second} \\
			$d_{\mathrm{w}}$ & \SI{1.93}{\meter} \\
			$d_{\mathrm{l}}$ & \SI{4.9}{\meter} \\
			\bottomrule
		\end{tabular}
		\qquad
		\begin{tabular}{c c}
			\toprule 
			Parameter & Value \\
			\midrule
			$l_{\mathrm{r}}$ & \SI{1.72}{\meter} \\
			$l_{\mathrm{f}}$ & \SI{1.25}{\meter} \\
			$v_{\mathrm{max}}$ & \SI{85}{\meter\per\second} \\
			$r_{\mathrm{min}}$ & \SI{1}{\meter} \\
			$\delta_{\mathrm{b},\max}$ & \SI{0.43}{\radian} \\
			$\omega_{\mathrm{b},\max}$ & \SI{0.39}{\radian\per\second} \\
			&\\
			\bottomrule
		\end{tabular}
		\label{tab:parameters}
	\end{center}
\end{table}

%% file: tables/evaluation_parameters.tex
\begin{table}[t]
	\caption{Evaluation and Training Parameters}
	\begin{center}
		\begin{tabular}{c c c c}
			\toprule 
			Parameter & Evaluation & Training \#1 & Training \#2 \\
			\midrule
			$s_\mathrm{b, init}$ & $\left\{20, 22, \dots, 100\right\}$ \si{\meter} & $\left[20, 100\right] \si{\meter}$ & $\left[20, 100\right]$ \si{\meter} \\
			$n_\mathrm{b, init}$ & $\left\{-6, -4, \dots, 6\right\}$ \si{\meter}   & $\left[-6, 6\right] \si{\meter}$ 	& $\left[-6, 6\right]$ \si{\meter} \\
			$v_\mathrm{init}$    & \SI{50}{\meter\per\second} 					   & \SI{50}{\meter\per\second} 		& \SI{50}{\meter\per\second} \\
			$s_{\mathrm{d}}$     & $\left\{40, 60, \dots, 140\right\}$             & \num{80} 							& $\left\{\num{40}, \num{80}, \num{120}\right\}$ \\
			$k_{\mathrm{p}}$     & \num{0.05} 									   & \num{0.05}							& \num{0.05} \\
			$k_{\mathrm{d}}$     & \num{0.6} 									   & \num{0.6} 							& \num{0.6} \\
			$k_{\mathrm{n}}$     & \num{1.0} 									   & \num{1.0} 							& \num{1.0} \\
			\bottomrule
		\end{tabular}
		\label{tab:evaluation_training_parameters}
	\end{center}
\end{table}

%% file: tables/cost_parameters.tex
\begin{table}[t]
	\caption{Cost Parameters for Conventional Approach}
	\begin{center}
		\begin{tabular}{c c c c c c c}
			\toprule 
			                  & \multicolumn{2}{c}{Small Ellipse} & \multicolumn{2}{c}{Medium Ellipse} & \multicolumn{2}{c}{Large Ellipse} \\
			                    \cmidrule(rl){2-3}                  \cmidrule(rl){4-5}                   \cmidrule(rl){6-7}
			Parameter         & \acs{CH}         & \acs{CLP}      & \acs{CH}        & \acs{CLP}        & \acs{CH}        & \acs{CLP}       \\
			\midrule
			$p_{\mathrm{s}}$  & \num{0.08}       & \num{0.08}     & \num{0.02}      & \num{0.02}       & \num{0.01}      & \num{0.01}      \\
			$p_{\mathrm{n}}$  & \num{0.5}        & \num{0.5}      & \num{0.18}      & \num{0.18}       & \num{0.1}       & \num{0.1}       \\
			$w_{\mathrm{pr}}$ & \num{5000}       & \num{5000}     & \num{5000}      & \num{5000}       & \num{5000}      & \num{5000}      \\
			$w_{\mathrm{n}}$  & \num{0.08}       & \num{0.0}      & \num{0.0}       & \num{0.72}       & \num{0.36}      & \num{0.8}       \\
			$w_{\mathrm{v}}$  & \num{0.28}       & \num{0.04}     & \num{0.08}      & \num{1.0}        & \num{0.24}      & \num{0.28}      \\
			\bottomrule
		\end{tabular}
		\label{tab:cost_parameters}
	\end{center}
\end{table}

%% file: figures/conventional_evaluation.tex
\begin{tikzpicture}

\definecolor{chocolate22711434}{RGB}{227,114,34}
\definecolor{darkcyan0101189}{RGB}{0,101,189}
\definecolor{darkgoldenrod1621730}{RGB}{162,173,0}
\definecolor{darkgray176}{RGB}{176,176,176}
\definecolor{lightgray204}{RGB}{204,204,204}

\begin{axis}[
height=\figH,
legend cell align={left},
legend style={
  fill opacity=0.8,
  draw opacity=1,
  text opacity=1,
  at={(0.97,0.03)},
  anchor=south east,
  draw=lightgray204
},
tick align=outside,
tick pos=left,
width=\figW,
x grid style={darkgray176},
xlabel={\(\displaystyle s_{\mathrm{d}}\)},
xmajorgrids,
xmin=35, xmax=145,
xtick style={color=black},
y grid style={darkgray176},
ylabel near ticks,
ylabel={success rate in \si{\percent}},
ymajorgrids,
ymin=-5, ymax=105,
ytick distance=25,
ytick style={color=black}
]
\addplot [very thick, darkcyan0101189, mark=*, mark size=1.5, mark options={solid}]
table {%
40 0
60 98.2578430175781
80 100
100 100
120 100
140 100
};
\addlegendentry{small (\acs{CH})}
\addplot [very thick, darkcyan0101189, dashed, mark=*, mark size=1.5, mark options={solid}]
table {%
40 0
60 86.7595825195312
80 98.6062698364258
100 100
120 100
140 100
};
\addlegendentry{small (\acs{CLP})}
\addplot [very thick, chocolate22711434, mark=triangle*, mark size=1.5, mark options={solid}]
table {%
40 29.6167259216309
60 75.2613220214844
80 95.1219482421875
100 95.8188171386719
120 94.0766525268555
140 99.3031387329102
};
\addlegendentry{med. (\acs{CH})}
\addplot [very thick, chocolate22711434, dashed, mark=triangle*, mark size=1.5, mark options={solid}]
table {%
40 0
60 19.1637630462646
80 86.0627212524414
100 96.8641128540039
120 99.6515655517578
140 100
};
\addlegendentry{med. (\acs{CLP})}
\addplot [very thick, darkgoldenrod1621730, mark=square*, mark size=1.5, mark options={solid}]
table {%
40 21.9512195587158
60 69.686408996582
80 81.8815307617188
100 97.9094085693359
120 97.5609741210938
140 94.0766525268555
};
\addlegendentry{large (\acs{CH})}
\addplot [very thick, darkgoldenrod1621730, dashed, mark=square*, mark size=1.5, mark options={solid}]
table {%
40 8.71080112457275
60 10.1045303344727
80 17.4216022491455
100 81.8815307617188
120 98.2578430175781
140 99.3031387329102
};
\addlegendentry{large (\acs{CLP})}
\end{axis}

\end{tikzpicture}

%% file: figures/rl_based_evaluation.tex
\begin{tikzpicture}

\definecolor{chocolate22711434}{RGB}{227,114,34}
\definecolor{darkcyan0101189}{RGB}{0,101,189}
\definecolor{darkgray176}{RGB}{176,176,176}
\definecolor{lightgray204}{RGB}{204,204,204}

\begin{axis}[
height=\figH,
legend cell align={left},
legend style={
  fill opacity=0.8,
  draw opacity=1,
  text opacity=1,
  at={(0.97,0.03)},
  anchor=south east,
  draw=lightgray204
},
tick align=outside,
tick pos=left,
width=\figW,
x grid style={darkgray176},
xlabel={\(\displaystyle s_{\mathrm{d}}\)},
xmajorgrids,
xmin=35, xmax=145,
xtick style={color=black},
y grid style={darkgray176},
ylabel near ticks,
ylabel={success rate in \si{\percent}},
ymajorgrids,
ymin=65, ymax=105,
ytick style={color=black}
]
\addplot [very thick, darkcyan0101189, mark=*, mark size=1.5, mark options={solid}]
table {%
40 73.1707305908203
60 99.6515655517578
80 99.3031387329102
100 98.954704284668
120 98.6062698364258
140 98.2578430175781
};
\addlegendentry{training \#1}
\addplot [very thick, chocolate22711434, dashed, mark=triangle*, mark size=1.5, mark options={solid}]
table {%
40 91.9860610961914
60 99.6515655517578
80 99.6515655517578
100 99.3031387329102
120 99.3031387329102
140 98.6062698364258
};
\addlegendentry{training \#2}
\end{axis}

\end{tikzpicture}

%% file: figures/safety_layer_evaluation.tex
\begin{tikzpicture}

\definecolor{chocolate22711434}{RGB}{227,114,34}
\definecolor{darkcyan0101189}{RGB}{0,101,189}
\definecolor{darkgray176}{RGB}{176,176,176}
\definecolor{lightgray204}{RGB}{204,204,204}

\begin{axis}[
height=\figH,
legend cell align={left},
legend columns=2,
legend style={
  fill opacity=0.8,
  draw opacity=1,
  text opacity=1,
  at={(0.5,0.5)},
  anchor=center,
  draw=lightgray204
},
tick align=outside,
tick pos=left,
width=\figW,
x grid style={darkgray176},
xlabel={\(\displaystyle s_{\mathrm{d}}\)},
xmajorgrids,
xmin=35, xmax=145,
xtick style={color=black},
y grid style={darkgray176},
ylabel near ticks,
ylabel={rate in \si{\percent}},
ymajorgrids,
ymin=-5, ymax=105,
ytick distance=25,
ytick style={color=black}
]
\addplot [very thick, chocolate22711434, mark=triangle*, mark size=1.5, mark options={solid}]
table {%
40 74.2160263061523
60 88.1533126831055
80 86.0627212524414
100 87.1080169677734
120 82.5783996582031
140 86.7595825195312
};
\addlegendentry{success w/o \ac{SL}}
\addplot [very thick, chocolate22711434, dashed, mark=triangle*, mark size=1.5, mark options={solid}]
table {%
40 18.8153305053711
60 11.8466901779175
80 13.2404184341431
100 12.5435543060303
120 16.3763065338135
140 11.8466901779175
};
\addlegendentry{infeasibility w/o \ac{SL}}
\addplot [very thick, darkcyan0101189, mark=*, mark size=1.5, mark options={solid}]
table {%
40 91.9860610961914
60 98.2578430175781
80 98.954704284668
100 99.3031387329102
120 99.3031387329102
140 98.6062698364258
};
\addlegendentry{success w/ \ac{SL}}
\addplot [very thick, darkcyan0101189, dashed, mark=*, mark size=1.5, mark options={solid}]
table {%
40 0
60 0
80 0
100 0
120 0
140 0
};
\addlegendentry{infeasibility w/ \ac{SL}}
\end{axis}

\end{tikzpicture}

%% file: chapters/conclusion_outlook.tex
\section{Conclusion and Outlook}
\label{sec:conclusion_outlook}
This paper investigates an existing conventional sampling-based and a novel \ac{RL}-based trajectory planning approach for autonomous racing in an interactive blocking scenario. The conventional approach, which relies on the prediction of the opposing vehicle, achieves high success rates for less aggressive blocking behaviors, as the prediction reflects the opposing behavior well. However, with more aggressive blocking behavior and thus a stronger interaction, the prediction loses validity, and the success rate decreases significantly. In contrast, the \ac{RL}-based approach proposed here explicitly exploits the interactive blocking behavior, allowing planning of maneuvers that are not realizable with the conventional approach. This results in high success rates even for the most aggressive blocking behavior. Here, different training configurations demonstrate the importance of various opponent behaviors for generalization. We further propose an \ac{SL} that generates a sub-optimal but valid trajectory in the event of an infeasible trajectory generated by the \ac{RL} agent. In the examined scenario, the \ac{SL} entirely counteracts infeasibility issues and increases the success rates.

In this article, we focus on a simple scenario to demonstrate and motivate the basic principle. However, a fundamental problem with \ac{RL}-based approaches is the lack of generalization to scenarios not included in training. If the \ac{RL} agent used here is evaluated against an opposing vehicle that behaves differently from the blocking behavior described in this article, the success rate may decrease. Future research could, therefore, involve a generalization to more complex scenarios and race tracks. More than one opposing vehicle with respective complex behaviors could be included to force more demanding and realistic maneuvers. This would require dealing with a dynamically changing state space size or, alternatively, a fundamental change of the state space, e.g., using fixed-size 2D LiDAR sensor data. In addition, instead of a flat straight track, complex circuits and their 3D effects could be investigated, requiring an extension of the state space by track information. Furthermore, multi-agent \ac{RL}, i.e., the simultaneous training of multiple agents in a single environment, is a possible extension. This avoids assumptions about the behavior of opposing vehicles, which are currently necessary. Finally, advanced network architectures such as recurrent or graph neural networks can be considered to better represent the complex interactions between agents.